# A Reply to "Is Complexity An Illusion?"

Gabriel Simmons

gsimmons@ucdavis.edu

28 Oct 2024

## Abstract

The paper "Is Complexity an Illusion?" (Bennett, 2024) provides a formalism for complexity, learning, inference, and generalization, and introduces a formal definition for a "policy". This reply shows that correct policies do not exist for a simple task of supervised multi-class classification, via mathematical proof and exhaustive search. Implications of this result are discussed, as well as possible responses and amendments to the theory.

## In appreciation:

The paper "Is Complexity an Illusion?" (Bennett, 2024) provides a formalism to talk about complexity, learning, inference, and generalization, among other things. These concepts are all at the heart of the field of artificial intelligence. The paper offers an exciting perspective on abstraction, namely that the notion of complexity is relative to the choice of an abstraction layer. As a result, (absolute, observer-independent) complexity is an illusion. The paper also claims to show that in choosing a proxy, one should seek to maximize policy weakness rather than minimizing complexity[1]. As I understand it, this paper extends and challenges the discourse around AIXI and Legg-Hutter intelligence (Legg & Hutter, 2007), (Hutter, 2005; Leike & Hutter, 2015), familiar concepts to those interested in AGI theory. Bennett questions the centrality of "complexity" and provides insight as to why Occam's Razor holds true so much of the time, despite there being no clear justification for why this should be so.

## Correct policies may not exist for simple tasks

**Criticism:** The criticism presented here is that simple tasks may be constructed for which no correct policies exist. This is shown through an enumeration of the possible policies in the next section. I also provide a Python program that iterates over all possible policies to prove the claim exhaustively.

**Why does this matter?:** Many of the exciting claims of Bennett's paper are about policies and how to find them. While the theory is abstract, I am optimistic that it has implications for practical AI systems. I would hope that "policies" in the abstract sense bear some resemblance to "policies" in commonplace AI systems like reinforcement learning agents or supervised classifiers. Likewise for tasks - I would hope that the theory could be used to reason about commonplace tasks like image classification or game playing.

I believe that the example presented in this reply is isomorphic to a very simple task with the following form:

| input | output | |
|---|---|---|
|  |  | is red |
|  |  | is blue |

---

[1] Any mistakes in interpretation are my own. This description is meant to emphasize the exciting takeaways of the paper, and may be imprecise.

I hope that machine learning practitioners would agree that correct policies for such tasks can be found. Such a policy might be described as follows:

> If the input is red, predict "red". If the input is blue, predict "blue".

I find that a policy like the above is surprisingly difficult to define in the framework of Bennett's paper. For a task isomorphic to the task above, no "correct policy" can be found, following the definition for "correct policy" given in the paper.

This highlights a gap between Bennett's theory and practical AI systems. I hope that this gap can be bridged, as I am optimistic about – and excited by – the theory's direction and implications.

## Definitions

I refer the reader to (Bennett 2024) Section 2 ("The Formalism") for definitions of the following terms: *state, environment, declarative program, fact, vocabulary, formal language, statement, completion, extension of a statement, extension of a set of statements, task, policy,* and *correct policy*.

In brief (all definitions from Bennett 2024):

- A set $\Phi$ is assumed, whose elements are called *states*.
- $f \subseteq \Phi$ is a declarative program. A declarative program is set of states.[2]
- $P = \mathcal{P}\Phi$ is the set of all possible declarative programs, the power set of $\Phi$.
- A subset $v \subseteq P$ is a *vocabulary*.
- A vocabulary implies a *formal language* $L_v = \{l \subseteq v : \cap l \neq \emptyset\}$, whose members $l \in L_v$ are called *statements*.
- A completion of a statement $x$ is any statement $y$ such that $x \subseteq y$.
- The extension of a statement is the set of all completions. For statement $x$ in $L_v$, its extension $E_x = \{y \in L_v : x \subseteq y\}$
- The extension of a set of statements $X \subseteq L_v$ is $E_X = \cup_{x \in X} E_x$, the union of the extensions of the statements in $X$.
- A v-task (hereafter, "task") $\alpha$ is a pair $\langle I_\alpha, O_\alpha \rangle$ where $I_\alpha \subset L_v$ are the *inputs* and $O_\alpha \subset E_{I_\alpha}$ are the *correct outputs*.
- A policy $\pi$ is a statement in $L_v$.
- A policy $\pi$ is a correct policy for task $\alpha$ (written $\pi \in \Pi_\alpha$) if and only if $E_{I_\alpha} \cap E_\pi = O_\alpha$

## Result

I will now present a simple task for which no correct policy exists.

Consider an environment with 5 states.

Recall that a declarative program is a set of states. The notation `01111` is used to denote a program that returns true in all states except state 1. The notation `10111` is used to denote a program that returns true in all states except state 2, and so on.

Consider the following four declarative programs:

$$f_1 = 01111 \quad f_2 = 10111$$
$$f_3 = 11011 \quad f_4 = 11101$$

We could use an alternative notation like the below:

---

[2]Readers can also think of a declarative program as a function that maps from states to boolean values, and consider the set of states for which the function returns true as a way to identify the declarative program.

$$f_1 = \{2, 3, 4, 5\} \quad f_2 = \{1, 3, 4, 5\}$$
$$f_3 = \{1, 2, 4, 5\} \quad f_4 = \{1, 2, 3, 5\}$$

Let the vocabulary be $v = \{f_1, f_2, f_3, f_4\}$. Note that all programs include state 5, so any subsets of the vocabulary will have a non-empty intersection (state 5 being the contents of the intersection).

Consider the following task $\alpha$:

$$I_\alpha = \{\{f_1\}, \{f_2\}\}$$
$$O_\alpha = \{\{f_1, f_3\}, \{f_2, f_4\}\}$$

**Claim: No correct policy exists for the task $\alpha$ above.**

**Proof:** This can be done by considering all policies. Recall that a policy is a statement in the vocabulary, and the set of all statements is the power set of the vocabulary.

There are $2^4 = 16$ possible statements in the language $L_v$, one of which is the empty set. These are the set of possible policies.

We can write down the extension of the inputs:

$$E_{I_\alpha} = E_{\{f_1\}} \cup E_{\{f_2\}}$$
$$= \{\{f_1\}, \{f_2\}, \{f_1, f_2\}, \{f_1, f_3\}, \{f_1, f_4\}, \{f_2, f_3\}, \{f_2, f_4\},$$
$$\{f_1, f_2, f_3\}, \{f_1, f_3, f_4\}, \{f_1, f_2, f_4\}, \{f_2, f_3, f_4\}, \{f_1, f_2, f_3, f_4\}\}$$

To check if a policy is correct, we are interested in determining if $E_{I_\alpha} \cap E_\pi = O_\alpha$. The extension of a policy is the set of statements in the language for which the policy is a subset. Since we know we will take an intersection with $E_{I_\alpha}$, we can ignore statements not appearing in $E_{I_\alpha}$. In other words, we can view the policy as selecting from the set of statements in $E_{I_\alpha}$. This selection must equal $O_\alpha$ for the policy to be correct.

At this point we can observe that a correct policy for $\alpha$ cannot contain more than 2 elements. All statements in the extension of a policy with 3 or more elements will themselves have 3 or more elements, since the extension of a statement is the set of all its completions. Since our set of correct outputs consists of statements of length 2, a policy with 3 or more elements cannot select statements in $O_\alpha$. So we must only consider policies of length 0, 1, or 2 as candidates for correct policies. We proceed to check each case.

For the length 0 case with $\pi = \emptyset$, $E_{I_\alpha} \cap E_\emptyset = E_{I_\alpha} \cap L_v = E_{I_\alpha}$. See Appendix 3. Since $O_\alpha \subset E_{I_\alpha}$ (the outputs are strictly a subset of the extension of the inputs), $E_{I_\alpha} \neq O_\alpha$. So $\pi = \emptyset$ is not a correct policy. (It may be surprising that empty policies *do* belong to $L_v$, for any choice of $v$. See Appendix 2.)

Next we consider an example of the length 1 case. For $\pi = \{f_1\}$, the policy overlaps with $E_{I_\alpha}$ on those entries containing $f_1$, namely $E_{I_\alpha} \cap E_\pi = \{f_1\}, \{f_1, f_2\}, \{f_1, f_3\}, \{f_1, f_4\}, \{f_1, f_2, f_3\}, \{f_1, f_3, f_4\}, \{f_1, f_2, f_4\}, \{f_1, f_2, f_3, f_4\}$. This is not equal to $O_\alpha$, so $\pi$ is not a correct policy.

Likewise, $\pi = \{f_2\}$ selects 8 statements from $E_{I_\alpha}$, $\pi = \{f_3\}$ selects 6 statements, and $\pi = \{f_4\}$ selects 6 statements. None of these policies are correct, since $O_\alpha$ consists of only 2 statements.

This leaves only the length 2 case. For the length 2 case, there are 6 possible policies to consider. Each of these length-2 policies has a length-3 completion in its extension that is also found in $E_{I_\alpha}$. Since $O_\alpha$ does not contain any length-3 statements, the intersection $E_{I_\alpha} \cap E_\pi$ cannot be equal to $O_\alpha$ in any of the 6 cases.

As the correct policy must have length 0, 1, or 2, and no correct policy exists among these, no correct policy exists for the task $\alpha$. ∎

# Discussion

## Relevance to real-world tasks

First, I would like to explain the isomorphism between the colored box task on the first page, and the task $\alpha$ used in the proof. My hope in this section is that readers who view the task $\alpha$ in the proof above as contrived will see it as a natural way to represent a common machine learning task.

Let $f_1$ denote something like "the box is actually red", or "a camera recorded high signal intensity in its red channel". Let $f_2$ denote something like "a typical human would think the image is red". Similar logic applies for $f_2$ (the box is actually blue), and $f_4$ (a typical human would think the image is blue). My rationale for choosing this task is that it is a simplied version of the task that an image classifier performs. That is, mapping low-level information about pixels to higher-level concepts like "red" and "blue"[3]. The colored box task can be seen as an extremely simplified case with a single pixel.

In the example task $\alpha$, there are some declarative programs that serve as "labels". These are the programs only appearing in $O_\alpha$ and not in $I_\alpha$. Likewise there are some programs that serve as "features" - these are the programs appearing in $I_\alpha$. I have adopted a convention where the task inputs are "labeled" by appending a "label program" ($f_3$ or $f_4$) in the example to the set of features to obtain the output. To me this seems the most natural way to represent a classification problem.

## Possible responses

Readers might respond that the task setup is not the correct, conventional, or expected way that Bennett's framework would be applied. Bennett's work is recent, so I imagine a convention has not yet been established. Part of my aim in writing this reply is to encourage work in this direction.

Readers might contend that the example is too simple. One could argue that it is too much of a simplification to use only 1 program for each input – typical machine learning problems use tens, hundreds, or thousands of features. I would argue in response that a framework that can handle such complex problems should also be able to handle simpler ones, and conjecture (without having verified this) that the results would hold even if $f_1$ and $f_2$ were replaced with larger sets of programs.

## Implications for the theory

As it stands, it seems difficult to represent multi-class classification tasks, or any tasks requiring conditional behavior in Bennett's framework. It is straightforward to train a machine learning classifier to represent a policy like "label all images that contain wheels as `car`, unless there are also train tracks, in which case the correct label is `train`". It seems natural that "image-is-a-car" and "image-is-a-train" should be part of the vocabulary, and that the policy should be able to select from these. But since the policy is a statement, it can include only one of these "label" facts. Including both would not allow the policy to select from correct outputs with only one of these labels.

Perhaps it makes sense to view a "task" in Bennett's framework as a binary classification task. Machine learning practitioners are aware that a multi-class classification task can be decomposed into a series of binary classification tasks. In this case, perhaps a "policy" for the multiclass task would be better represented by a set of statements, each one performing a binary classification. The definition for inference from Bennett's paper would need almost no adjustment, since the "extension" operation accommodates both statements and sets of statements.

Perhaps my intuition that labels can be represented as programs is false – perhaps there should not be an "image-is-a-train" fact. In this case, it seems important for the theory to provide an alternative way to represent "labels" in the practical machine learning sense.

---

[3] or "dog" and "cat", or "malignant" and "benign", or any of the other tasks that are popular in computer vision.

# Acknowledgements

Thanks to Fangzhou Li and Vladislav Savinov for helpful discussion and review of this reply.

# Appendix 1

This Python program exhaustively checks all non-empty policies for correctness with respect to the task $\alpha$. Empty policies *can* be included in $L_v$ (see Appendix 2), but are not checked here since it is trivial to show that the empty policy is not a correct policy (see "Result" section).

```python
from itertools import chain, combinations
from pprint import pprint
from typing import Collection

show_work = True  # set to True to see the work for each policy

def powerset(iterable):
        # https://stackoverflow.com/questions/1482308/how-to-get-all-subsets-of-a-set-powerset
    # modified so that the empty set is not included
    # this is a convenience. The empty set _is_ included in languages L_v in general,
    # but is trivially not a correct policy and makes the programming less convenient
    "powerset([1,2,3]) --> (1,) (2,) (3,) (1,2) (1,3) (2,3) (1,2,3)"
    s = list(iterable)
    return [
        frozenset(x)
        for x in chain.from_iterable(combinations(s, r) for r in range(0, len(s) + 1))
    ]

def print_statement(statement: frozenset):
    pprint(sorted(statement, key=len))

assert bool(int("0")) == False
assert bool(int("1")) == True

def non_null_intersection(programs: Collection[str]) -> bool:
    if len(programs) == 0:
        return False

    length_to_check = max(map(len, programs))

    for i in range(length_to_check):
        if all(bool(int(p[i])) for p in programs):
            return True

    return False

assert non_null_intersection(["01", "11"])
assert not non_null_intersection(["01", "10"])

assert non_null_intersection(["011", "111"])
assert not non_null_intersection(["011", "100"])

def extension_of_statement(statement: frozenset, L_v: frozenset) -> frozenset:
    if statement not in L_v:
```

```python
        raise ValueError

    return frozenset({y for y in L_v if statement.issubset(y)})

def extension_of_set_of_statements(
    statements: frozenset[frozenset], L_v: frozenset
) -> frozenset:
    return frozenset(
        chain.from_iterable(extension_of_statement(s, L_v) for s in statements)
    )

# Let there be 5 states
# A declarative program is a set of states
# The set of all possible programs is the power set of the set of all states

# We are allowed to take a subset of all possible programs to form a vocabulary.
# let us take the following subset of programs:

# We can identify four programs with non-null intersection
f_1 = "01111"  # program f_1 includes all states except state 1
f_2 = "10111"  # program f_2 includes all states except state 2
f_3 = "11011"  # program f_3 includes all states except state 3
f_4 = "11101"  # program f_4 includes all states except state 4
# all programs f_1, f_2, f_3, f_4 include state 5

# In this string notation, a program includes state i if the i-th character is 1,
# and excludes it if the i-th character is 0

v = {
    f_1,
    f_2,
    f_3,
    f_4,
}
if show_work:
    print("v:")
    pprint(v)
    print("")

assert non_null_intersection({f_1, f_2, f_3, f_4})

# From our vocabulary, we define a formal language - this is the set of all
# sets of facts that are realized by the environment, meaning there is at least one state
# that is shared by all programs in the set
L_v = frozenset({l for l in powerset(v) if non_null_intersection(l)})

if show_work:
    print("L_v:")
    print_statement(L_v)
    print("")

# now we can test the extension of a statement, and extension of a set of statements:
```

```python
statement1 = frozenset({f_1, f_2})

assert extension_of_statement(statement1, L_v) == frozenset(
    {
        frozenset({f_1, f_2}),
        frozenset({f_1, f_2, f_3}),
        frozenset({f_1, f_2, f_4}),
        frozenset({f_1, f_2, f_3, f_4}),
    }
)

statement2 = frozenset({f_2, f_3})

assert extension_of_statement(statement2, L_v) == frozenset(
    {
        frozenset({f_2, f_3}),
        frozenset({f_2, f_3, f_1}),
        frozenset({f_2, f_3, f_4}),
        frozenset({f_2, f_3, f_1, f_4}),
    }
)

assert (
    extension_of_set_of_statements(frozenset({statement1, statement2}), L_v)
    == extension_of_statement(statement1, L_v).union(
        extension_of_statement(statement2, L_v)
    )
    == frozenset(
        {
            frozenset({f_1, f_2}),
            frozenset({f_1, f_2, f_3}),
            frozenset({f_1, f_2, f_4}),
            frozenset({f_1, f_2, f_3, f_4}),
            frozenset({f_2, f_3}),
            frozenset({f_2, f_3, f_1}),
            frozenset({f_2, f_3, f_4}),
            frozenset({f_2, f_3, f_1, f_4}),
        }
    )
)

# For programming convenience, this script ignores the empty set.
assert len(powerset(v)) == 2 ** len(v) == 16
assert len(L_v) == 15

# Now we will show that no correct policy exists for a simple task that is compatible
with Definition 3 from Bennett 2024
# we can construct a task <I_alpha, O_alpha> as follows,
# where I_alpha is the set of task inputs,
# and O_alpha is the set of correct outputs:

I_alpha = frozenset(
    {
        frozenset({f_1}),
        frozenset({f_2}),
```

```python
        }
    )

    # frozenset is used to make the inner sets hashable, so the larger set O can be defined
    O_alpha = frozenset(
        {
            frozenset({f_1, f_3}),
            frozenset({f_2, f_4}),
        }
    )

    # This task <I_alpha, O_alpha> satisfies the requirements in Defintion 3, namely that
    # - I is a proper subset of L_v
    assert I_alpha.issubset(L_v) and I_alpha != L_v
    # - O_alpha is a subset of the extension of I_alpha
    assert O_alpha.issubset(
        extension_of_set_of_statements(I_alpha, L_v)
    ) and O_alpha != extension_of_set_of_statements(I_alpha, L_v)

    # Is there a correct policy for the task <I_alpha, O_alpha>?
    # According to the paper, a policy pi is correct if (E_i sect E_pi) = O_alpha

    # A policy is a statement pi in L_v
    # Let's iterate over all possible policies

    correct_policy_exists = False

    for i, policy in enumerate(L_v):
        policy_extension = extension_of_statement(policy, L_v)  # E_pi
        input_extension = extension_of_set_of_statements(I_alpha, L_v)  # E_i
        intersection_of_extensions = policy_extension.intersection(input_extension)

        if show_work:
            print(f"\n ----------------- policy {i}: -----------------")
            print(f"policy {i}:")
            print_statement(policy)
            print(f"\npolicy {i} extension:")
            print_statement(policy_extension)
            print(f"\ninput extension:")
            print_statement(input_extension)
            print(f"\nintersection of extensions:")
            print_statement(intersection_of_extensions)
            print(f"\nO_alpha:")
            print_statement(O_alpha)

        if intersection_of_extensions == O_alpha:
            print("correct policy")
            correct_policy_exists = True

            print(f"correct policy is policy {i}:")
            print_statement(policy)

    print("\nAll policies checked")

    if not correct_policy_exists:
        print("no correct policy exists")
```

## Appendix 2

**Claim:** Given some set of states $\Phi$, for any vocabulary $v \subseteq \mathcal{P}\Phi$, $\emptyset \in L_v$.

**Proof:**

We know that $L_v = \{l \subseteq v : \cap l \neq \emptyset\}$ (definition of a formal language, Bennett).

Thus $\emptyset \in L_v$ iff $\emptyset \subseteq v$ and $\cap \emptyset \neq \emptyset$.

For any choice of $v$, $\emptyset \subseteq v$ is true. The empty set is a subset of all sets[4].

When $\emptyset$ is the empty set of subsets of some set $X$, $\cap \emptyset = X$[5,6]. In our case $\emptyset$ is the empty set of declarative programs, the programs being subsets of $\Phi$. Thus for any choice of $v$, $\cap \emptyset = \Phi \neq \emptyset$ is true.

Thus $\emptyset \in L_v$ is true for all $v$. ∎

## Appendix 3

**Claim:** For some set of states $\Phi$, for any vocabulary $v \subseteq \mathcal{P}\Phi$, $E_\emptyset = L_v$.

**Proof:**

$E_x = \{y \in L_v : x \subseteq y\}$ (definition of extension, Bennett)

$E_\emptyset = \{y \in L_v : \emptyset \subseteq y\}$

$\emptyset \subseteq y$ for all $y \in L_v$ (the empty set is a subset of all sets)

$E_\emptyset = \{y \in L_v\} = L_v$ ∎

---

[4]https://proofwiki.org/wiki/Empty_Set_is_Subset_of_All_Sets
[5]https://en.wikipedia.org/wiki/Intersection_(set_theory)#Nullary_intersection,
[6]P.R. Halmos, Naive Set Theory (1960), Chapter 5